\documentclass[letterpaper]{article} 
\usepackage{aaai25}  
\usepackage{times}  
\usepackage{helvet}  
\usepackage{courier}  
\usepackage[hyphens]{url}  
\usepackage{graphicx} 
\usepackage{amsmath}
\usepackage{tabularx}
\usepackage{makecell}
\usepackage{multirow}
\usepackage{adjustbox}
\usepackage{amsfonts}  
\usepackage{amssymb}
\usepackage{colortbl}
\usepackage{natbib}  
\usepackage{caption} 
\usepackage{algorithm}
\usepackage{algorithmic}
\newcommand{\eg}{\textit{e.g.}}
\usepackage{newfloat}
\usepackage{listings}
\DeclareCaptionStyle{ruled}{labelfont=normalfont,labelsep=colon,strut=off} 
\floatstyle{ruled}
\newfloat{listing}{tb}{lst}{}
\floatname{listing}{Listing}
\title{{S$^3$-Mamba}: Small-Size-Sensitive Mamba for Lesion Segmentation}
\author{
    Gui Wang\textsuperscript{\rm 1,2},
    Yuexiang Li\textsuperscript{\rm 3},
    Wenting Chen\textsuperscript{\rm 4},
    Meidan Ding\textsuperscript{\rm 1},
    Wooi Ping Cheah\textsuperscript{\rm 2},
    Rong Qu\textsuperscript{\rm 5},\\
    Jianfeng Ren\textsuperscript{\rm 2}\footnotemark[1],
    Linlin Shen\textsuperscript{\rm 1}\footnotemark[1]
}
\affiliations{
 
    \textsuperscript{\rm 1}Computer Vision Institute, School of Computer Science \& Software Engineering, Shenzhen University, Shenzhen, China\\
    \textsuperscript{\rm 2}School of Computer Science, University of Nottingham Ningbo China, Ningbo, Zhejiang, China\\
    \textsuperscript{\rm 3}Medical AI ReSearch (MARS) Group, University Engineering Research Center of Digital Medicine and Healthcare, Guangxi Medical University, Nanning, Guangxi, China\\
    \textsuperscript{\rm 4}Department of Electrical Engineering, City University of Hong Kong, Kowloon, Hong Kong\\
    \textsuperscript{\rm 5}School of Computer Science, University of Nottingham, Nottingham, United Kingdom\\
    scxgw1@nottingham.edu.cn, yuexiang.li@sr.gxmu.edu.cn, wentichen7-c@my.cityu.edu.hk, dingmeidan2023@email.szu.edu.cn, wooi-ping.cheah@nottingham.edu.cn, rong.qu@nottingham.ac.uk, jianfeng.ren@nottingham.edu.cn, llshen@szu.edu.cn\\
    
}

\usepackage{bibentry}

\begin{document}

\maketitle
\footnotetext[1]{Corresponding Authors: Linlin Shen and Jianfeng Ren.}

\begin{abstract}
Small lesions play a critical role in early disease diagnosis and intervention of severe infections. Popular models often face challenges in segmenting small lesions, as it occupies only a minor portion of an image, while down\_sampling operations may inevitably lose focus on local features of small lesions. To tackle the challenges, we propose a {\bf S}mall-{\bf S}ize-{\bf S}ensitive {\bf Mamba} ({\bf S$^3$-Mamba}), which promotes the sensitivity to small lesions across three dimensions: channel, spatial, and training strategy. Specifically, an Enhanced Visual State Space block is designed to focus on small lesions through multiple residual connections to preserve local features, and selectively amplify important details while suppressing irrelevant ones through channel-wise attention. A Tensor-based Cross-feature Multi-scale Attention is designed to integrate input image features and intermediate-layer features with edge features and exploit the attentive support of features across multiple scales, thereby retaining spatial details of small lesions at various granularities. Finally, we introduce a novel regularized curriculum learning to automatically assess lesion size and sample difficulty, and gradually focus from easy samples to hard ones like small lesions. Extensive experiments on three medical image segmentation datasets show the superiority of our S$^3$-Mamba, especially in segmenting small lesions. Our code is available at https://github.com/ErinWang2023/S3-Mamba.
\end{abstract}

\section{Introduction}

\begin{figure}[t]
\centering
\includegraphics[width=1\columnwidth]{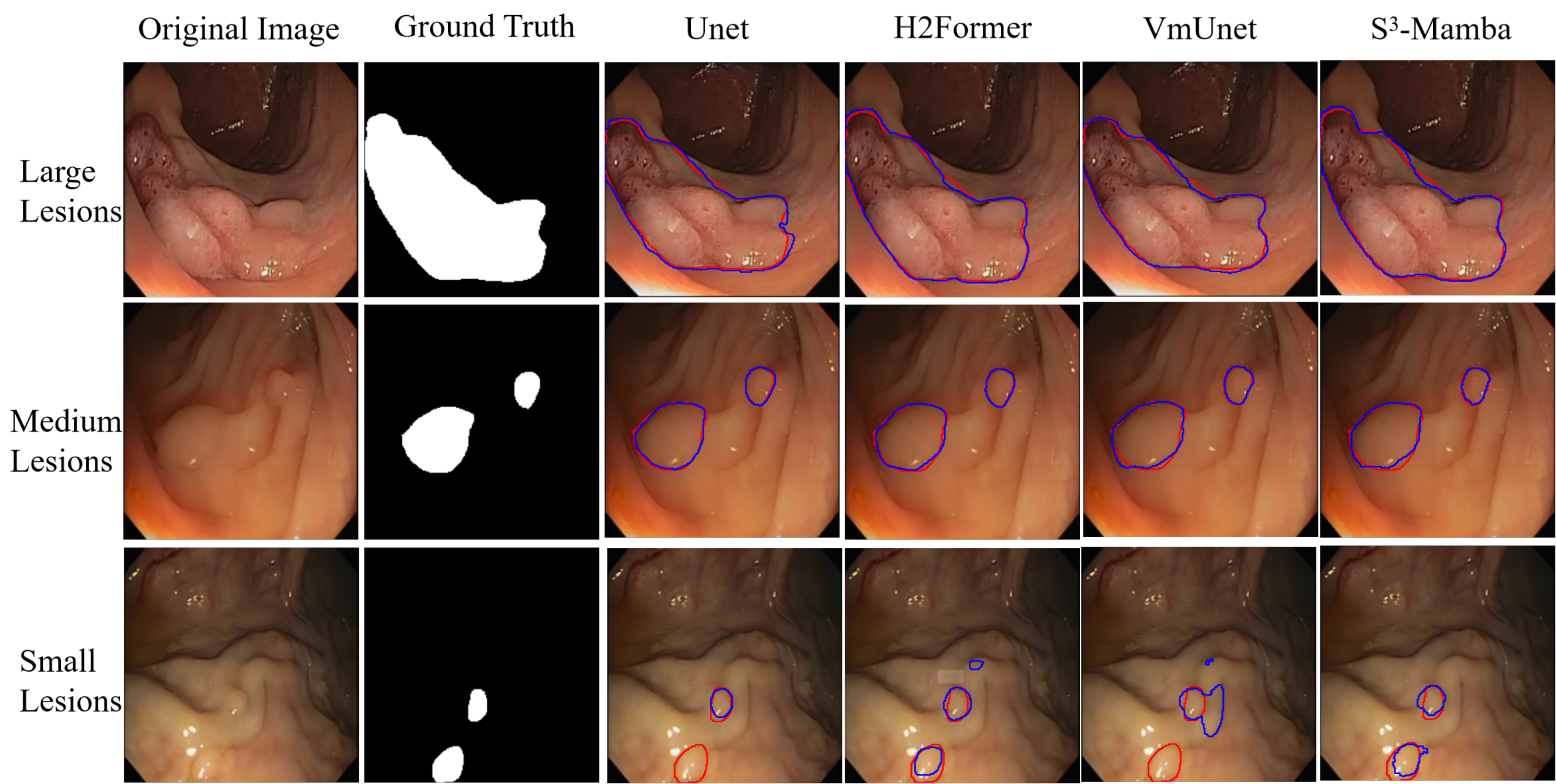} 
\caption{Segmentation results of CNN-based Unet~\cite{Unet}, transformere-based H2Former~\cite{H2former}, Mamba-based VmUnet~\cite{2024vmunet}, and the proposed S$^3$-Mamba for lesions of varying sizes. All models perform well in segmenting large and medium lesions, but S$^3$-Mamba exhibits superior performance in segmenting small lesions, capturing fine details and contours of small lesions with higher accuracy. 
}
\label{fig1}
\end{figure}

As the earliest indicators of severe diseases, small lesions play a pivotal role in clinical outcomes, where timely and precise identification can dramatically improve patient prognosis \cite{review2023medical}. These lesions, despite their small size, often signal the onset of significant health issues, making their segmentation essential for effective treatment planning \cite{de2023convolutional, Nature}. Many images from different modalities in medical imaging \cite{greenwald2022whole, luo2024lesion} contain numerous lesions that occupy less than 10\% of the image area \cite{duering2023neuroimaging, polyp}, \eg, early tumors, micrometastases in lymph nodes, and small vascular abnormalities. 

Existing models for medical image segmentation such as U-Net \cite{Unet}, ViTSeg \cite{du2023avit}, and MedT \cite{MedT, SwinUNETR} have made significant strides in segmenting large objects in medical images. However, these models often struggle with the segmentation of small lesions. Fig. \ref{fig1} shows the results of several models for segmenting lesions. It is evident that small lesions are inaccurately segmented due to their subtle textures and morphology. A key challenge arises from the common down-sampling operations in network architectures, which unintentionally eliminates crucial local features associated with small lesions. As these lesions only occupy a small portion of the image, they may be overlooked or inadequately represented in the segmentation output, requiring more advanced techniques to handle small lesions. 

To tackle the challenges, we propose a \textbf{S}mall-\textbf{S}ize-\textbf{S}ensitive
\textbf{Mamba} (\textbf{S$^3$-Mamba}) to effectively and efficiently segment small lesions. In view of the balance performance of Mamba \cite{gu2023mamba} between accuracy and speed, we choose Mamba as the baseline model. The proposed S$^3$-Mamba enhances the segmentation of small lesions through three novel techniques: an \textbf{En}hanced \textbf{V}isual \textbf{S}tate \textbf{S}pace (\textbf{EnVSS}) block to selectively amplify key features, an \textbf{T}ensor-based \textbf{C}ross-feature \textbf{M}ulti-scale \textbf{A}ttention (\textbf{TCMA}) to integrate and preserve critical spatial details, and a novel curriculum learning strategy to adapt the training process based on lesion size and sample difficulty. 

More specifically, the proposed EnVSS block aims to tackle the problem of losing local features of small lesions during down-sampling through learnable channel weights and residual connections. Specifically, we insert the \textbf{En}hanced \textbf{C}hannel \textbf{F}eature Block (EnCFBlock) into the Visual State Space block \cite{Vmamba}, which integrates global context information through squeeze, excitation and scale operations to generate channel statistics and scale channel features, so that contribution of each channel is automatically adjusted and fine-tuned based on its relevance to small lesion features. In addition, two residual connections are applied to preserve important lesion fine details by amplifying crucial features while downplaying less relevant ones.  
The proposed EnVSS block enhances the model’s ability to capture and emphasize the subtle characteristics of small lesions that are often overlooked in existing models. 

The TCMA is designed to retain the spatial characteristics of small lesions across multiple granularities, exploiting the attentive support from feature maps of multiple scales to enhance the sensitivity to small lesions. The novel tensor-based attention provides an effective mechanism to not only integrate multi-modality features at multiple scales but also dynamically adjust the attention based on the interplay between small lesion areas and their surrounding background context. Unlike existing methods that often treat each feature scale independently or focus solely on the lesion \cite{attention2, attentionchannel}, the proposed TCMA uniquely emphasizes the relationship between the lesion and the larger surrounding regions at multiple granularities. This dual focus well preserves the spatial integrity of small lesions while simultaneously using the broader context to ensure that critical small lesions are not overlooked.  

Traditional random sampling strategies \cite{singh2021order} often prioritize the accurate segmentation of larger lesions due to their dominance in the image, while neglecting small lesions. Although some workaround techniques such as data augmentation \cite{dataaugmentation} have been explored to replicate small lesions to improve their representation, the bias towards larger objects remains a challenge. 
Inspired by \cite{wang2021clsurvey}, we propose a novel regularized curriculum learning strategy to dynamically adjust the training process based on lesion size and sample difficulty. More specifically, a Difficulty Measurer is designed to assess the the difficulty of samples, initialize sample weights based on lesion size, and dynamically update them according to loss values during training. A Training Scheduler with regularization constraints is designed to gradually shift the focus from simpler samples to more challenging samples such as small lesions as training converges.  
This strategy enables the model to prioritize more challenging cases and adaptively focus on small lesion segmentation.

Our main contributions can be summarized as follows. 
1)~The proposed EnVSSBlock dynamically fine-tunes the contribution of each channel based on its relevance to small lesions through channel-wise attention and effectively preserves crucial lesion fine details through two residual connections.  
2)~The proposed TCMA employs a novel tensor-based multi-level attention strategy to integrate multi-modality features across scales and dynamically adjust the attention based on the interaction between small lesions and their surroundings, preserving small lesion integrity while leveraging surrounding context to better segment small lesions. 
3)~The proposed regularized curriculum learning assesses the sample difficulty through a Difficulty Measurer and gradual shifts from easier samples to more challenging ones such as small lesions, through a Training Scheduler.

\begin{figure*}[t]
\centering
\includegraphics[width=1\textwidth]{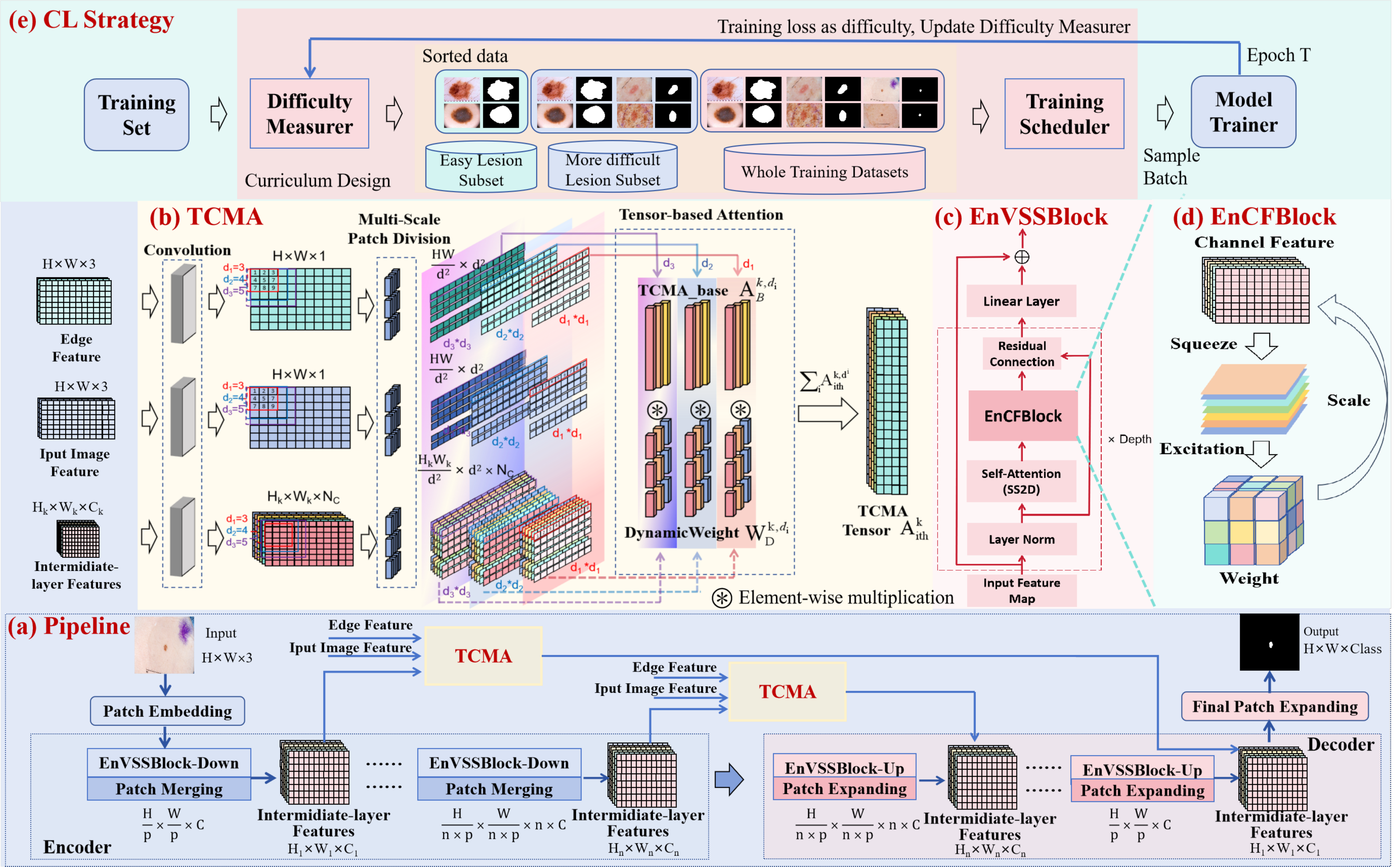} 
\caption{(a) Overview of the proposed S$^3$-Mamba with \textit{TCMA} and \textit{EnVSSBlock}. (b) Detailed architecture of the TCMA, where input image features, intermediate-layer features, and edge features are divided into patches of three different sizes. A tensor-based attention derives the dynamic weights of these patches of three different scales, exploits their interaction, and utilizes the TCMA features to modulate the features at decoder layers. (c) The detailed structure of the EnVSSBlock, which explicitly evaluates and adaptively adjusts the channel weights to enhance small lesion feature representation and preserves fine details through residual connections. (d) Detailed structure of the Enhanced Channel Feature Block (\textit{EnCFBlock}), which enhances the feature interaction through channel-wise interaction. (e) The architecture of the regularized curriculum learning strategy. 
}
\label{fig2}
\end{figure*}

\section{Related Work}
\noindent\textbf{Medical Image Segmentation.}
Existing models for medical image segmentation can be broadly categorized into four groups based on their architecture: 1) Models based on convolutional neural networks (CNNs)\cite{Unet, vnet, chen2020tr, xu2024hacdr, 10122681, luo2023mvcinn, li2018skin}; 2) Transformer-based models such as \cite{UNETR++, MedT, H2former}; 3) Hybrid models combining CNN and Transformer architectures\cite{swin-unet, SwinUNETR, chen2024bichen2022tw, MedFormer}; and 4) State space models, like VmUNet \cite{2024vmunet, UMamba, segmamba}. Although these models have made significant strides, small lesion segmentation remains challenging due to the down-sampling operations that commonly exist in these models, which may overlook small targets. 
In literature, there are some attempts for small object detection, \eg, increase the input image size to generate high-resolution feature maps \cite{dataaugmentation, wang2024generative},  feature enhancement \cite{smallobject}, feature pyramids \cite{gao2024augmented, mei2023pyramid} and tuning loss functions \cite{miao2023sc, chen2022tw}. These methods offer valuable insights for segmenting small lesions. However the unique characteristics of medical images, \eg, low contrast, complex anatomical backgrounds, and diverse imaging modalities, often complicate segmentation tasks and increase the difficulty of accurately segmenting small lesions, highlighting the urgent need for tailored models to address these challenges.

Very recently, Mamba models \cite{visionMamba, Vmamba} have gained traction, especially for medical image segmentation. U-Mamba \cite{UMamba} extends the Mamba to the U-Net architecture, aiming to capture long-range dependencies using hybrid CNN-SSM (State Space Model) blocks. 
VM-UNET \cite{2024vmunet} incorporates the Visual State Space (VSS) block to capture contextual information, and its asymmetric encoder-decoder structure demonstrates the capabilities of SSM for segmentation. These Mamba-based models predominantly emphasize long-range dependencies and contextual information, while sacrificing the accuracy of segmenting small lesions. To bridge this gap, we propose an enhanced VSS block that reduces the computational complexity while improving the segmentation accuracy of small lesions.

\noindent\textbf{Curriculum Learning.} 
Curriculum learning \cite{wang2021clsurvey} was initially introduced as a training strategy to gradually increase sample complexity during training, mimicking the structured learning of humans and improving model generalization and convergence speed \cite{CL}. Recent innovations in Curriculum Learning (CL) include dynamic CL~\cite{wang2019dynamic}, which adjusts task difficulty based on training progress to enhance generalization; MTL-based CL~\cite{kong2021adaptive}, which organizes tasks hierarchically to improve shared learning across tasks; RL-driven CL~\cite{zhang2024scale}, which utilizes reinforcement learning to dynamically adjust the task difficulty for flexible and efficient learning. These methods primarily focus on dynamically adapting the task complexity to facilitate progressive learning. In this paper, a novel curriculum learning strategy with regularization is designed to help focus on more accurately segmenting challenging small lesions. The proposed approach improves upon existing methods by introducing a Difficulty Measurer to evaluate the sample difficulty based on lesion size and model training loss and implementing a Training Scheduler with regularization constraints to balance sample weights and prevent the model from overfocusing on specific samples. 

\section{Proposed Method}
\subsection{Overview of Proposed S$^3$-Mamba} 
As shown in Fig. \ref{fig2}(a), our S$^3$-Mamba is built upon a U-shape encoder-decoder architecture, including an Enhanced Visual State Space Block (\textit{EnVSSBlock}) and a Tensor-based Cross-feature Multi-scale Attention (\textit{TCMA}). 
A patch embedding operation \cite{Vmamba} is first applied to the original image to generate initial feature maps. The encoder comprises a series of \textit{EnVSSBlocks\_Down} and patch merging operations, progressively capturing global dependencies, extracting small lesion features, and compressing spatial dimensions. 
The TCMA makes use of edge features, input image features, and intermediate-layer features, and divides them into patches of three distinct sizes,  aiming to enrich the diversity of extracted features and focus on small lesions. Then, novel tensor-based attention exploits the attentive supports from these multi-modal mult-scale patch features and merges them to produce the final TCMA features, which are applied to tune the features from decoder layers. The decoder consists of a series of \textit{EnVSSBlocks\_Up} and patch-expanding operations that progressively reconstruct the spatial dimensions while enhancing small lesion features.  
To further focus on small lesions, we introduce a regularized curriculum learning strategy, including a Difficulty Measurer to assess sample difficulty and a Training Scheduler to focus on segmenting challenge targets like small lesions. 

\subsection{Enhanced Visual State Space Block}
\textit{VSSBlock} is the core module of VMamba \cite{Vmamba}, capable of effectively capturing rich contextual information through deep convolution and \textit{SS2D}, making it perform well in segmentation tasks. 
However, the \textit{VSSBlock} has limitations in retaining fine details due to the local aggregation of deep convolution and the indiscriminate handling of channel features, causing these fine features to be excessively smoothed or overshadowed by more prominent background information. To tackle these challenges, we propose the \textit{EnVSSBlock} to better preserve image fine details. 

The improvements over \textit{VSSBlock}  are three-fold. 
1)~We remove the \textit{DWConv} layer in \textit{VSSBlock}, thereby avoiding excessive smoothing and reducing propagation loss of information, enabling the model to retain more detailed features.  
2)~We introduce two residual connections as shown in Fig.~\ref{fig2}(c), one from the input directly to the output that helps maintain the integrity of the initial input features, preventing the information loss in a deep network, and the other highlighting the important features to small lesions while preserving local details.   
 
3)~An \textit{EnCFBlock} is added to enhance channel features through adaptive reweighting, suppressing irrelevant background context, and highlighting the key features to more accurately represent small lesions. 
These improvements collectively help the model focus on segmenting small lesions by preserving local details and adaptively highlighting the features relevant to small lesions. 

The \textit{EnCFBlock} exploits the channel-wise attention as in \cite{hu2018squeeze}. Specifically, a global average pooling is first applied to capture the global context of each channel; The squeezed features are then passed through two fully connected layers to further reduce the dimensionality and then restore it, followed by a Sigmoid activation function; These adaptive weights are then applied to the previous channel features, enhancing important features while suppressing less relevant ones; Finally, a residual connection is employed to to preserve locality. Hence, the \textit{EnCFBlock} enhances the feature interaction through channel-wise attention to highlight local features relevant to small lesions. 

\subsection{Tensor-based Cross-feature Multi-scale Attention} 
Existing attention mechanisms such as Soft Attention \cite{omeroglu2023novel}, Hard Attention \cite{jegham2023deep}, Self-Attention \cite{cao2023masactrl}, and Multi-Head Attention \cite{wang2024emat} have demonstrated their power in various tasks, but they exhibit significant limitations when applied to small lesion segmentation, as they are not originally designed to effectively handle multi-scale information, nor can they fully integrate features of different types. What's more, attention weights are often fixed after computation \cite{luvembe2023dual}, nor adaptive to varying contexts. 

To tackle the challenges, we propose a Tensor-based Cross-feature Multi-scale Attention (TCMA)  to enhance the model's ability to segment small lesions in three aspects. 
1)~The TCMA dynamically tunes the feature representations based on multi-feature multi-scale tensors through a novel tensor-based attention mechanism. The tensors encapsulate the input image features, intermediate prediction features, and edge features at multiple scales. 
The joint utilization of these features could help exploit the information relevant to segment small lesions in a wider spatial context and a broader range of modalities.  
2)~The novel tensor-based attention provides an effective mechanism to exploit the attentive supports from features of different modalities at different scales. This unique approach allows for simultaneous interaction between spatial, categorical (class-related), and edge features, enabling the model to dynamically focus on the most relevant features for precisely segmenting small lesions.  
3)~Finally, the integration of TCMA tensors from low-level features in encoder layers with high-level abstract features in decoder layers provides an effective mechanism to exploit both sets of features, which captures the key features of small lesions in both global context and fine details in local neighborhoods. 
As a result, the proposed TCMA leads to more precise segmentation of small lesions.

\noindent\textbf{Generation of Multi-Feature Multi-Scale Tensors.} 
We first construct a set of pyramid features, including the input image features embedded using a $7\times7$ convolutional layer and a $3\times3$ convolutional layer, the edge features extracted using a Sobel operator and embedded with a $3\times3$ convolutional layer, and intermediate prediction features. More specifically, 
Given the \( i \)-th level feature maps \(\mathsf{F}_i \in \mathbb{R}^{H_i \times W_i \times C_i} \), a 1×1 convolutional layer $\mathcal{F}_{1\times1}$ with \( N_c \) filters converts $\mathsf{F}_i$ into per-category prediction features as 
$\mathsf{R}_{i} = \mathcal{F}_{1\times1}(\mathsf{F}_i)$, where \(\mathsf{R}_{i} \in \mathbb{R}^{H_i \times W_i \times N_c} \) and \( N_c \) is the number of categories to predict. We hypothesize that \(\mathsf{R}_{i}^k \in \mathbb{R}^{H_i \times W_i \times 1} \) (for \( 1 \leq k \leq N_c \)) contains only the information associated with the \( k \)-th category.
We then divide the feature pyramid \( \mathsf{R} \in \mathbb{R}^{H \times W \times C} \) into patches of three different scales as,
\begin{equation}
\mathsf{P}^{d_i} = \mathcal{F}_\pi(\mathsf{R}, d_i),
\end{equation}
where \( \mathcal{F}_\pi \) is the image partition operation,  \( d_i \) represents the size of patches, and \( \mathsf{P}^{d_i} \in \mathbb{R}^{\frac{H_i W_i}{d_i^2} \times d_i^2 \times N_c} \). 
This multi-feature multi-scale tensor provides a detailed and comprehensive feature representation that enhances the model's ability to accurately identify and segment small lesions.

\noindent\textbf{Tensor-based Attention.}  
While the tensor $\mathsf{P}^{d_i}$ encapsulates multi-feature multi-scale information, it is difficult to exploit the attentive support through the interaction among the tensor features. To achieve this, a novel tensor-based attention is designed to dynamically tune the weights of the tensor features. 
Specifically, an \texttt{einsum} operation is firstly adopted 
to exploit the relations between edge features, input image features, and the \( k \)-th category-related features as,  
\begin{equation}
\begin{split}
\mathsf{A}_{\text{B}}^{k,d_i} = \sum_{j=1}^{d_i^2} \mathsf{P}_{\text{O}}^{d_i}(b, i, j, c) \cdot 
\mathsf{P}_{\text{I}}^{k,d_i}(b, i, j, k) \cdot \mathsf{P}_{\text{E}}^{d_i}(b, i, j, e),
\end{split}
\end{equation}

where \(\mathsf{P}_{\text{O}}^{d_i}(b, i, j, c)\) represents the tensor value derived from the original input image features for the $b$-th sample in the batch, the $i$-th patch, the $j$-th pixel and the $c$-th channel; \(\mathsf{P}_{\text{I}}^{k,d_i}(b, i, j, k)\) represents the tensor value related to the \(k\)-th category from the intermediate prediction features; \(\mathsf{P}_{\text{E}}^{d_i}(b, i, j, e)\) represents the tensor value of edge features for the $e$-th channel. By summing over all pixel positions within the patch, we aggregate the information from all pixels of a patch into the TCMA base tensor $\mathsf{A}_{\text{B}}^{k,d_i}$. 

The element-wise multiplication between 
\(\mathsf{P}_{\text{O}}^{d_i}(b, i, j, c)\), \(\mathsf{P}_{\text{I}}^{k,d_i}(b, i, j, k)\), and \(\mathsf{P}_{\text{E}}^{d_i}(b, i, j, e)\) 
encapsulates the combined influence of the spatial, categorical (class-related), and edge features, which are crucial for small lesion segmentation. In particular, the spatial features provide location-based context and the general appearance of the lesion, the categorical features ensure the model focuses on the relevant class (\eg, small lesions), and the edge features help to accurately delineate the boundaries of the lesion, which is often challenging due to blurring or low contrast. 
By combining all of them, the model generates a more robust and accurate representation $\mathsf{A}_{\text{B}}^{k,d_i}$. 
\(\mathsf{P}_{\text{O}}^{d_i} \), \( \mathsf{P}_{\text{I}}^{k,d_i} \), and \( \mathsf{P}_{\text{E}}^{d_i} \) are then concatenated along the last dimension to form a comprehensive tensor
$\mathsf{A}_{\text{C}}^{k,d_i} \in \mathbb{R}^{B \times P \times d_i^2 \times (C + N_c + E)}$,
and processed through an MLP to compute dynamic weights,  
\begin{equation}
    \mathsf{W}_{\text{D}}^{k,d_i} = \mathcal{F}_\text{MLP}(\mathcal{F}_\text{Flatten}(\mathsf{A}_{\text{C}}^{k,d_i})) \in \mathbb{R}^{B \times P \times d_i^2 \times O},
\end{equation}
where $\mathcal{F}_\text{Flatten}(\cdot)$ is the flatten operation, $\mathcal{F}_\text{MLP}(\cdot)$ represents fully connected layers with batch normalization, ReLU activation, and sigmoid activation functions, and \( O \) is the output dimension. 
The weighted TCMA tensor is then obtained by aggregating over the last dimension of the channel as, 
\begin{equation}
    \mathsf{A}_{\text{ith}}^{k,d_i}= \sum \mathsf{A}_{\text{B}}^{k,d_i} \cdot \mathsf{W}_{\text{D}}^{k,d_i}. 
\end{equation}

The TCMA tensor \( \mathsf{A}_{\text{ith}}^k \) is obtained by integrating the tensors from different scales as $\mathsf{A}_{\text{ith}}^k = \sum_{i} \mathsf{A}_{\text{ith}}^{k,d_i}$, ensuring the model's robustness for different sizes of lesions.

The combination of different types of features helps more accurately segment small lesions, which are often subtle and difficult to distinguish from the background. The utilization of multi-scale features allows the model to capture both fine details and broader contextual information, critical for handling lesions of varying sizes. The resulting tensor \( \mathsf{A}_{\text{ith}}^k \) enables the model to dynamically adjust its focus on the most relevant features, leading to more precise segmentation. 

\noindent\textbf{Integration of TCMA Tensors with Decoder Features.}   

The obtained TCMA tensor \( \mathsf{A}_{\text{ith}}^k \) is then applied to modulate the output of each \textit{EnVSSLayer\_up}, enhancing the segmentation of small objects. 
Given a predicted mask \(\hat{\mathsf{M}} \in \mathbb{R}^{H \times W \times N_c}\) of an \textit{EnVSSLayer\_up} layer, it is first resized using fixed-size average pooling to match the dimensions of an intermediate feature map, and reshaped to \(\tilde{\mathsf{M}} \in \mathbb{R}^{H_i \times W_i \times N_c}\) to match the dimensions of TCMA tensor. Then, \( \mathsf{A}_{\text{ith}}^k \) modulates the decoder features as,
\begin{equation}
\hat{\mathsf{M}}_{\text{ith}}^k = \mathsf{A}_{\text{ith}}^k \otimes \tilde{\mathsf{M}}^k,
\end{equation}
where $\otimes$ denotes the \texttt{einsum} operation. This operation integrates the multi-modal multi-scale image fine details conveyed by the TCMA with the global context in the decoder features, refining the focus on small lesions in the network.

\begin{table*}[t]
\centering
\small
\renewcommand{\arraystretch}{1.2} 
\setlength{\tabcolsep}{3pt} 
\begin{tabular}{p{0.3cm} p{0.3cm}l c c c|c c c|c c c|c c c|c c c}
    \hline
    \multirow{2}{*}{} &\multirow{2}{*}{} & \multirow{2}{*}{\textbf{MODELS}} & \multicolumn{3}{c|}{\textbf{MIOU}} & \multicolumn{3}{c|}{\textbf{DSC}} & \multicolumn{3}{c|}{\textbf{ACC}} & \multicolumn{3}{c|}{\textbf{SPE}} & \multicolumn{3}{c}{\textbf{SEN}} \\
    \cline{4-18}
    & & & \textbf{S} & \textbf{M} & \multicolumn{1}{c|}{\textbf{L}} & \textbf{S} & \textbf{M} & \multicolumn{1}{c|}{\textbf{L}} & \textbf{S} & \textbf{M} & \multicolumn{1}{c|}{\textbf{L}} & \textbf{S} & \textbf{M} & \multicolumn{1}{c|}{\textbf{L}} & \textbf{S} & \textbf{M} & \textbf{L} \\
    \hline
    \multirow{10}{*}{\rotatebox{90}{\textbf{ISIC2018}}}
    & \multirow{1}{*}{\rotatebox{90}{\small CNN}}
    & UNet $_{\text{(MICCAI'15)}}$ & \small 64.36 & \small 79.31 & \small 83.60 & \small 77.78 & \small 88.46 & \small 91.07 & \small 98.16 & \small 97.13 & \small 92.65 & \small 98.30 & \small 97.99 & \small 96.01 & \small 94.37 & \small 90.86 & \small 88.11 \\
    & & UNETR++ $_{\text{(TMI'24)}}$ & \small 55.72 & \small 74.17 & \small 82.20 & \small 71.56 & \small 85.17 & \small 90.23 & \small 97.44 & \small 96.22 & \small 92.02 & \small 97.55 & \small 97.14 & \small 95.96 & \small 94.46 & \small 89.60 & \small 86.69 \\
    & \multirow{3}{*}{\rotatebox{90}{\small Trans}} 
    & MedT $_{\text{(MICCAI'21)}}$ & \small 56.19 & \small 73.19 & \small 81.57 & \small 71.93 & \small 84.52 & \small 89.85 & \small 97.52 & \small 96.04 & \small 91.62 & \small 97.67 & \small 96.95 & \small 94.91 & \small 33.00 & \small 89.38 & \small 87.18 \\
    & & H2Former $_{\text{(TMI'23)}}$ & \small 63.51 & \small 77.61 & \small 83.26 & \small 77.68 & \small 87.39 & \small 90.87 & \small 98.18 & \small 96.85 & \small 92.50 & \small 98.36 & \small 97.76 & \small 96.03 & \small 93.06 & \small 90.24 & \small 87.73 \\
    & & ViTSeg $_{\text{(MICCAI'23)}}$ & \small 55.12 & \small 73.83 & \small 80.50 & \small 71.07 & \small 84.94 & \small 89.19 & \small 97.51 & \small 96.18 & \small 91.20 & \small 97.79 & \small 97.16 & \small 95.47 & \small 89.76 & \small 89.05 & \small 85.42 \\
    &\multirow{3}{*}{\rotatebox{90}{\small Hybrid}}
    & SwinUnet $_{\text{(ECCV'22)}}$ & \small 62.89 & \small 79.36 & \small 81.38 & \small 77.21 & \small 88.49 & \small 89.74 & \small 98.14 & \small 97.19 & \small 91.67 & \small 98.34 & \small 98.31 & \small 96.14 & \small 92.53 & \small 89.10 & \small 85.62 \\
    & & SwinUNETR $_{\text{(CVPR'22)}}$ & \small 71.36 & \small 78.63 & \small 82.98 & \small 83.29 & \small 88.03 & \small 90.70 & \small 98.73 & \small 97.08 & \small 92.25 & \small 98..93 & \small 98.21 & \small 94.83 & \small 93.08 & \small 88.83 & \small 88.77 \\
    & & MedFormer $_{\text{(arXiv'23)}}$ & \small 60.78 & \small 77.00 & \small 84.52 & \small 75.60 & \small 87.00 & \small 91.61 & \small 97.91 & \small 96.64 & \small 92.68 & \small 98.01 & \small 97.17 & \small 93.74 & \small \textbf{95.01} & \small \textbf{92.84} & \small \textbf{91.66} \\
    &\multirow{2}{*}{\rotatebox{90}{Mam}}
    & VmUnet $_{\text{(MICCAI'24)}}$ & \small 64.04 & \small 78.90 & \small \textbf{84.67} & \small 78.07 & \small 88.21 & \small \textbf{91.70} & \small 98.21 & \small 97.03 & \small \textbf{93.07} & \small 98.37 & \small 97.76 & \small 95.42 & \small 93.67 & \small 91.75 & \small 89.91 \\
    & & \textbf{S$^3$-Mamba (Ours)} & \small \textbf{77.13} & \small \textbf{81.36} & \small 83.28 & \small \textbf{87.09} & \small \textbf{89.72} & \small 90.87 & \small \textbf{99.07} & \small \textbf{97.55} & \small 92.60 & \small \textbf{99.34} & \small \textbf{98.81} & \small \textbf{97.01} & \small 91.66 & \small 88.39 & \small 86.65 \\
    \hline
    \multirow{10}{*}{\rotatebox{90}{\textbf{CVC-ClinicDB}}} 
    &\multirow{1}{*}{\rotatebox{90}{\small CNN}}
    & UNet $_{\text{(MICCAI'15)}}$ & \small 69.17 & \small 78.13 & \small 79.87 & \small 81.78 & \small 87.72 & \small 88.81 & \small 99.02 & \small 98.52 & \small 96.88 & \small 99.40 & \small 99.33 & \small 99.02 & \small 84.80 & \small 86.19 & \small 84.44 \\
    & & UNETR++ $_{\text{(TMI'24)}}$ & \small 36.07 & \small 51.89 & \small 55.98 & \small 53.02 & \small 68.33 & \small 71.78 & \small 97.00 & \small 96.09 & \small 92.13 & \small 97.84 & \small 97.86 & \small 96.22 & \small 65.43 & \small 68.96 & \small 68.32 \\
    &\multirow{3}{*}{\rotatebox{90}{\small Trans}}
    & MedT $_{\text{(MICCAI'21)}}$ & \small 23.56 & \small 39.05 & \small 47.15 & \small 38.13 & \small 56.16 & \small 64.08 & \small 94.67 & \small 93.59 & \small 90.33 & \small 95.50 & \small 95.31 & \small 95.72 & \small 63.42 & \small 67.13 & \small 58.91 \\
    & & H2Former $_{\text{(TMI'23)}}$ & \small 65.82 & \small 81.02 & \small 86.61 & \small 79.39 & \small 89.52 & \small 92.82 & \small 98.85 & \small 98.73 & \small 97.93 & \small 99.21 & \small 99.41 & \small 99.11 & \small 85.40 & \small 88.41 & \small 91.12 \\
    & & ViTSeg $_{\text{(MICCAI'23)}}$ & \small 13.48 & \small 22.42 & \small 30.50 & \small 23.76 & \small 36.62 & \small 49.06 & \small 91.57 & \small 90.03 & \small 87.50 & \small 92.66 & \small 92.83 & \small 90.54 & \small 50.72 & \small 47.09 & \small 41.10 \\
    &\multirow{3}{*}{\rotatebox{90}{\small Hybrid}}
    & SwinUnet $_{\text{(ECCV'22)}}$ & \small 24.06 & \small 39.43 & \small 53.89 & \small 38.79 & \small 56.56 & \small 70.04 & \small 94.56 & \small 93.55 & \small 91.90 & \small 95.30 & \small 95.18 & \small 96.59 & \small 66.60 & \small 68.59 & \small 64.61 \\
    & & SwinUNETR $_{\text{(CVPR'22)}}$ & \small 49.89 & \small 65.05 & \small 71.33 & \small 66.57 & \small 78.82 & \small 83.27 & \small 98.07 & \small 97.36 & \small 95.52 & \small 98.70 & \small 98.48 & \small 98.87 & \small 74.22 & \small 80.21 & \small 76.03 \\
    & & MedFormer $_{\text{(arXiv'23)}}$ & \small 60.89 & \small 74.35 & \small 80.78 & \small 75.69 & \small 85.29 & \small 89.37 & \small 98.61 & \small 98.16 & \small 96.98 & \small 99.01 & \small 98.88 & \small 98.77 & \small 83.66 & \small 87.11 & \small 86.55 \\
    &\multirow{2}{*}{\rotatebox{90}{\small Mam}} 
    & VmUnet $_{\text{(MICCAI'24)}}$ & \small 58.26 & \small 82.24 & \small \textbf{87.94} & \small 73.63 & \small 90.25 & \small \textbf{93.58} & \small 98.37 & \small 98.79 & \small 98.14 & \small 98.66 & \small 99.36 & \small 99.14 & \small 87.73 & \small \textbf{91.83} & \small \textbf{92.34} \\
    & & \textbf{S$^3$-Mamba (Ours)} & \small \textbf{75.40} & \small \textbf{83.81} & \small 85.86 & \small \textbf{85.97} & \small \textbf{91.19} & \small 92.39 & \small \textbf{99.25} & \small \textbf{99.46} & \small \textbf{99.08} & \small \textbf{99.53} & \small \textbf{99.81} & \small \textbf{99.58} & \small \textbf{88.78} & \small 89.61 & \small 91.37 \\
    \hline
\end{tabular}
\caption{Performance comparison on ISIC2018 and CVC-ClinicDB datasets. \textbf{S}, \textbf{M} and \textbf{L} represent small, medium, and large lesions respectively. Our S$^3$-Mamba performs the best for most metrics on most settings, especially for small lesions. 
}
\label{table_comparison}
\end{table*}

\subsection{Regularized Curriculum Learning Strategy}

We further refine the training strategy to fully unlock the model's potential in segmenting small lesions through curriculum learning shown in Fig. \ref{fig2}(e). We establish a Difficulty Measurer based on the size of the lesions. During training, we dynamically update the Difficulty Measurer based on the loss values, assigning higher weights to samples with lower losses and vice versa. As training progresses and loss stabilizes, the Training Scheduler shifts focus to difficult samples with lower weights. 
Specifically, given a dataset $\{(\mathbf{x}_i, \mathbf{y}_i)\}_{i=1}^N$, where $\mathbf{x}_i$ represents features and $\mathbf{y}_i$ denotes labels for each instance $i$, the model $\mathcal{F}_\phi$, parameterized by $\phi$, generates a prediction $\mathcal{F}_\phi(\mathbf{x}_i)$ for each $\mathbf{x}_i$ and computes the corresponding loss $l_i = \mathcal{L}(\mathcal{F}_\phi(\mathbf{x}_i), \mathbf{y}_i)$, where $\mathcal{L(\cdot)}$ denotes the loss function. The primary objective is to minimize the empirical loss across the entire training set, 
\begin{equation}
\min_\phi \sum_{i=1}^N v_i l_i + g(\textbf{v}), 
\end{equation}

where $\mathbf{v} = [v_1, v_2, \ldots, v_N] \in [0, 1]^N$ represents the vector of weights for training samples, determined by the Difficulty Measurer. 
Different from traditional curriculum learning~\cite{wang2021clsurvey}, we introduce a novel regularization term 
$g(\textbf{v})$ to influence the selection of $v_i$ as, 
\begin{equation}
g(\mathbf{v}) = \lambda \sum_{i=1}^N \left( \frac{1}{\mathcal{F}_\text{rank}(l_i)} \right) + (1-\lambda) \sum_{i=1}^N v_i^2,
\end{equation}
where \( \lambda \) is a weighting factor to balance the influence of weight assignment using loss ranking and the strength of the weight squared sum regularization, and $\mathcal{F}_\text{rank}(\cdot)$ is a rank function. Both terms serve as regularization constraints, with the former to regularize the weight distribution, ensuring the model focuses on easier samples first and gradually transit to more difficult ones such as small lesions, and the latter to regularize the sample weights \(v_i\) to prevent excessively large weights, 
ensuring a balanced distribution of weights.

\section{Experimental Results}
\noindent\textbf{Datasets.} 
The ISIC2018 dataset~\cite{ISIC2018} contains 2,694 dermoscopy images specifically designed for lesion segmentation. The CVC-ClinicDB dataset~\cite{CVC}, a benchmark for colonoscopy image analysis, includes 612 high-resolution colonoscopy images with corresponding polyp annotations, focusing on polyp detection and segmentation. The in-house Lymph dataset consists of 5,344 multi-modal MRI images of prostate cancer lymph nodes, including 2,733 T2-weighted (T2WI) and 2,611 T1-weighted (T1WI) images, annotated by three skilled radiologists with cross-validation to ensure reliable ground truth.

We then analyze lesion sizes across the three datasets. 
We sort the lesion pixel distributions from smallest to largest in the ISIC2018 and CVC-ClinicDB datasets, dividing them into three groups: the smallest 30\% as small lesions, 30\%-60\%  as medium lesions, and 60\% and above as large lesions. We randomly select 30\% from each group to create three separate sets for model testing, each focusing on small, medium, and large lesions, respectively. The remaining 70\% samples are used for model training. The Lymph dataset predominantly consists of very small and uniformly distributed objects. We utilize 70\% for training and 30\% for testing without further subdivision.

\noindent\textbf{Compared Methods.} Nine state-of-the-art segmentation models are compared, falling into four groups based on their architecture. 1) Traditional CNN-based models such as UNet~\cite{Unet}; 2) Attention-based models, including UNETR++~\cite{UNETR++}, MedT~\cite{MedT}, H2Former~\cite{H2former}, and ViTSeg~\cite{du2023avit}; 3) Hybrid models combining CNN and Transformer architectures, \eg, SwinUNet~\cite{swin-unet}, SwinUNETR~\cite{SwinUNETR}, and MedFormer~\cite{MedFormer}; and 4) State space models such as VmUNet~\cite{2024vmunet}.

\noindent\textbf{Implementation Details.} The input image size is $256\times256$ pixels. Common data augmentation techniques \cite{garcea2023data} are applied to boost the performance. 
The backbone is initialized using VMamba-S pre-trained on ImageNet-1k. The AdamW optimizer is used with an initial learning rate of 0.0001 and a cosine scheduler. The batch size is 16. The maximum number of epochs is 600. Experiments are conducted on a Tesla V100 GPU with 32GB memory. 

\noindent\textbf{Evaluation Metrics.} Five evaluation metrics are adopted: Mean Intersection over Union (mIoU), Dice Similarity Coefficient (DSC), Accuracy (ACC), Specificity (SPE), and Sensitivity (SEN), same as in~\cite{hirling2024segmentation}.

\begin{table}[t]
\centering
\setlength{\tabcolsep}{3pt} 
\renewcommand{\arraystretch}{1.2} 
\fontsize{9pt}{9pt}\selectfont
\begin{tabular}{p{0.3cm}l c c c c c}
    \hline
     & \textbf{MODELS} & \textbf{MIOU} & \textbf{DSC} & \textbf{ACC} & \textbf{SPE} & \textbf{SEN} \\
    \hline
    \multirow{1}{*}{\rotatebox{90}{{\small CNN}}}
    & UNet $_{\text{(MICCAI'15)}}$ & 48.56 & 65.37 & 99.91 & 99.98 & 56.88 \\
    \cline{2-7}
    \multirow{4}{*}{\rotatebox{90}{{\small Trans}}}
    & UNETR++ $_{\text{(TMI'24)}}$ & 48.68 & 65.48 & 99.92 & 99.98 & 57.33 \\
    & MedT $_{\text{(MICCAI'21)}}$ & 23.77 & 38.41 & 99.86 & 99.95 & 32.68 \\
    & H2Former $_{\text{(TMI'23)}}$ & 47.52 & 64.42 & 99.92 & 99.98 & 54.88 \\
    & ViTSeg $_{\text{(MICCAI'23)}}$ & 14.94 & 26.00 & 99.79 & 99.89 & 27.12 \\
    \cline{2-7}
    \multirow{3}{*}{\rotatebox{90}{{\small Hybrid}}}
    & SwinUnet $_{\text{(ECCV'22)}}$ & 19.64 & 32.83 & 99.85 & 99.95 & 26.51 \\
    & SwinUNETR $_{\text{(CVPR'22)}}$ & 45.03 & 62.09 & 99.91 & 99.97 & 54.02 \\
    & MedFormer $_{\text{(arXiv'23)}}$ & 55.25 & 71.18 & 99.92 & 99.97 & 68.78 \\
    \cline{2-7}
    \multirow{2}{*}{\rotatebox{90}{{\small Mam}}}
    & VmUnet $_{\text{(MICCAI'24)}}$ & 52.27 & 68.66 & 99.92 & 99.97 & 62.44 \\
    & \textbf{S$^3$-Mamba (Ours)} & \textbf{61.19} & \textbf{75.93} & \textbf{99.94} & \textbf{99.97} & \textbf{75.18} \\
    \hline
\end{tabular}
\caption{Performance comparison on the Lymph dataset.}
\label{table_lymph}
\end{table}

\begin{figure}[t]
\centering
\includegraphics[width=1\linewidth]{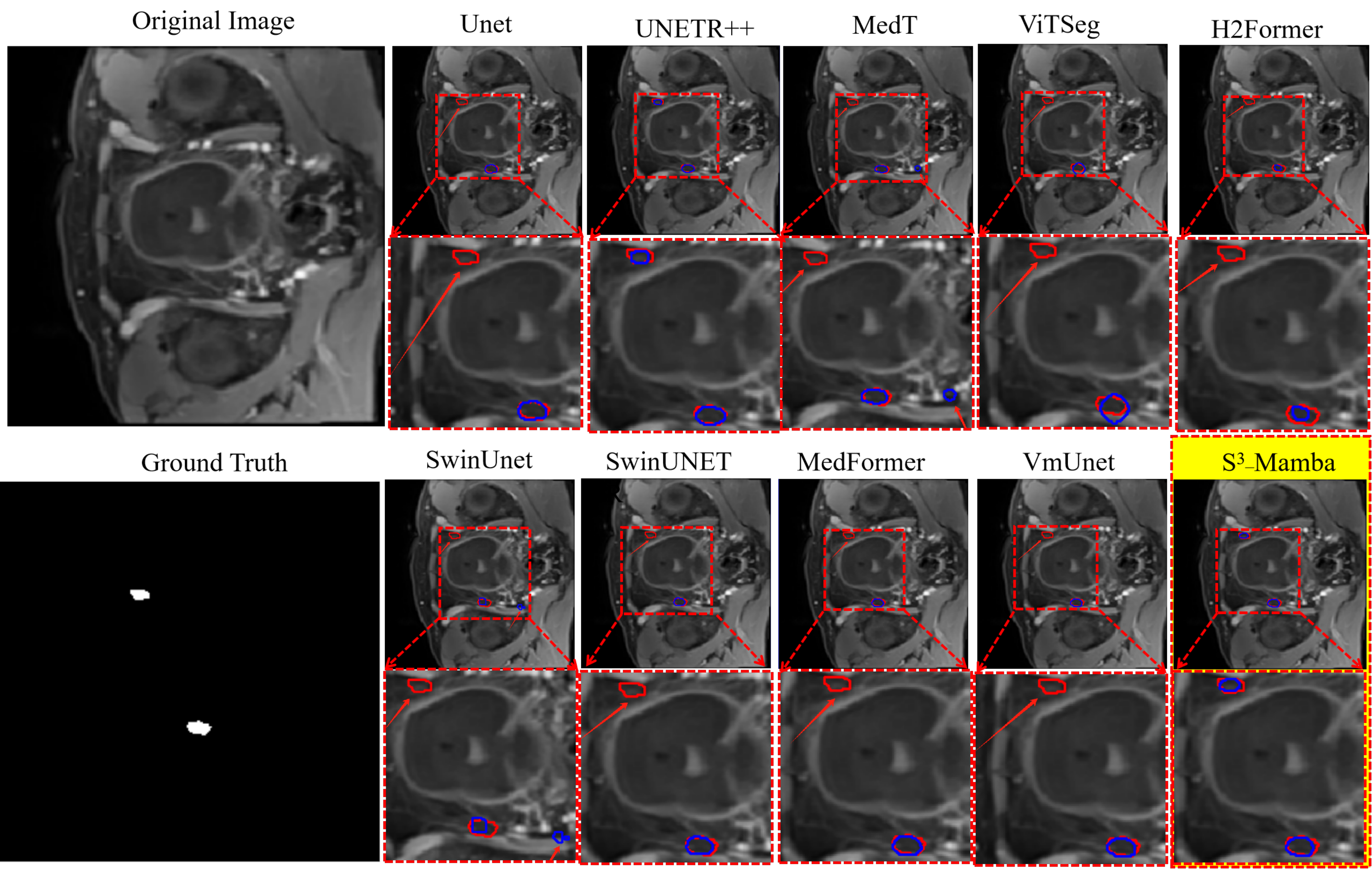} 
\caption{Segmentation results on the Lymph dataset. Red outlines represent the ground truth segmentation masks, and the blue marks indicate the model predictions.}

\label{figlymph}
\end{figure}

\subsection{Comparison with State-of-the-Art Methods}
\noindent\textbf{ISIC2018 Dataset.} 
The comparison results on the ISIC2018 dataset are summarized in Table \ref{table_comparison}. 
The following can be observed. 
1)~Our S$^3$-Mamba achieves the highest mIoU of 77.13\% and 81.36\% for small and medium lesions respectively, and a competitive mIoU of 83.28\% for large lesions. This robustness across lesion sizes is attributed to the EnVSS block's dynamic fine-tuning of channel contributions and the TCMA's multi-level attention strategy, both of which enhance the model's ability to preserve small lesion details while leveraging the surrounding context. 
2)~The VmUnet \cite{2024vmunet}, a predecessor of the Mamba series, excels in large lesion segmentation with a mIoU of 84.67\% and a DSC of 91.70\%, but it is less effective with small and medium lesions. This suggests that the VmUnet prioritizes broader spatial context, potentially at the cost of fine-grained details necessary for smaller lesions. The enhancements in our S$^3$-Mamba, particularly in attention mechanisms, likely contribute to its improved performance across different lesion sizes. 3)~UNet \cite{Unet} achieves an mIoU of 64.36\%, 79.31\%, 83.60\% for small, medium and large lesions respectively. While it is a foundational model in medical image segmentation, it lags behind newer models, particularly on small lesions, highlighting the need for ongoing innovation in segmentation techniques. 
4)~Transformer-based models like MedT achieves an mIoU of 56.19\%, 73.19\%, and 81.57\% and H2Former achieves an mIoU of 63.51\%, 77.61\%, and 83.26\% respectively for small, medium and large lesions. They exhibit varied performance, with H2Former particularly strong in small lesion segmentation. This suggests that transformers are highly dependent on their specific configuration and integration. 

\noindent\textbf{CVC-ClinicDB Dataset.} 
The following can be observed on the CVC-ClinicDB dataset. 
1)~Some models such as UNETR++~\cite{UNETR++}, MedT~\cite{MedT} and ViTSeg~\cite{du2023avit} experience a significant performance drop on this dataset compared to their performance on the ISIC2018 dataset, possibly due to the generally small lesions in this dataset, with a lesion pixel ratio ranging from 0.44\% to 50.17\% only. In addition, the ISIC2018 dataset utilizes high-resolution dermoscopic images, while the CVC-ClinicDB dataset consists of colonoscopic images where the visual characteristics of lesions differ.
2)~In contrast, our S$^3$-Mamba continues to excel on the CVC-ClinicDB dataset, because of its optimization for small lesion segmentation, \eg, the EnVSS block dynamically adjusts channel contributions, and the multi-level tensor-based attention integrates features across scales, preserving lesion integrity while leveraging the background information.
3)~While the proposed S$^3$-Mamba exhibits strong overall performance, its sensitivity (SEN) is not the highest across both datasets. The slightly lower sensitivity could be due to the model's balanced focus on both sensitivity and specificity.

\noindent\textbf{Private Lymph Dataset.} 
The comparison results for ultra-small lesions on the private prostate lymph dataset are summarized in Table \ref{table_lymph}. Our S$^3$-Mamba outperforms all the models, achieving an mIoU of 61.19\%, DSC of 75.93\%, accuracy of 99.94\%, and sensitivity of 75.18\%, respectively. While the performance of all models decreases on the Lymph dataset  compared to that on the ISIC2018 and CVC-ClinicDB datasets, our S$^3$-Mamba still leads by about 5\% in several metrics. 
Fig.~\ref{figlymph} shows sample segmentation results, particularly focusing on small lesions. S$^3$-Mamba demonstrates a closer alignment with the ground truth and produces fewer false positives compared to other models, as evident in the zoomed-in sections of the image. 
These observations demonstrate that S$^3$-Mamba’s design innovations significantly contribute to its leading performance on the Lymph dataset, particularly in small lesion segmentation. 

\subsection{Ablation Study of Key Components} 

We ablate the key components of the proposed S$^3$-Mamba model using the Lymph and ISIC2018 datasets. The baseline model is trained without the EnCFBlock, TCMA, or curriculum learning (CL). We then assess the impact of adding these components individually and in combination. Specifically, we evaluate the baseline with EnCFBlock (+EnCF), TCMA (+TCMA), and CL (+CL), as well as their combinations, such as EnCFBlock + TCMA (+TCMA+EnCF), EnCFBlock + CL (+EnCF+CL), and TCMA + CL (+TCMA+CL). The ablation results on the Lymph dataset and ISIC2018 dataset are summarized in Table~\ref{table_Ablation} and Table~\ref{table_AblatiISIC2018} respectively. 
\begin{table}[htpb]
\centering
\setlength{\tabcolsep}{3pt} 
\renewcommand{\arraystretch}{1.2} 
\fontsize{9pt}{9pt}\selectfont
\begin{tabular}{l c c c c c}
    \hline
    \textbf{Methods} &  \textbf{MIOU} &  \textbf{DSC} &  \textbf{ACC} &  \textbf{SPE} &  \textbf{SEN} \\
    \hline
    { Baseline} & {52.27} & {68.66} & {99.92} & {99.97} & {62.44} \\
    \hline
    {+EnCF} & {55.70} & {71.55} & {99.93} & {99.97} & {67.06} \\
    {+TCMA} & {55.66} & {71.51} & {99.93} & {99.97} & {68.67} \\
    {+CL} & {55.15} & {71.09}  & {99.92} & {99.97} & {67.82} \\
    \hline
    {\small +TCMA+EnCF} & {57.14} & {72.73} & {99.93} & {99.97} & {69.99} \\
    {+EnCF+CL} & {57.71} & {73.19}  & {99.93} & {99.96} & {72.82} \\
    {+TCMA+CL} & {59.62} & {74.70} & {99.93} & {99.97} & \textbf{77.60} \\
    \hline
    { \textbf{S$^3$-Mamba}} & { \textbf{61.19}} & { \textbf{75.93}} & { \textbf{99.94}} & { \textbf{99.97}} &  75.18 \\
     \hline
\end{tabular}
\caption{Ablation of key components on the Lymph dataset.} 
\label{table_Ablation}
\end{table}

As shown in Table \ref{table_Ablation}, the baseline achieves an mIoU of 52.27\%. Adding EnCFBlock increases the mIoU to 55.70\%, while TCMA results in a similar improvement (55.66\%). Combining TCMA and EnCFBlock (+TCMA+EnCF) further increases the mIoU to 57.14\%, indicating a synergistic effect between these two components. When curriculum learning (CL) is incorporated (+CL), the mIoU improves slightly to 55.15\%, suggesting its role in helping the model focus on small lesions. The combination of TCMA and CL (+TCMA+CL) further improves the mIoU to 59.62\%. Finally, the complete S$^3$-Mamba model, achieves the highest performance across all metrics, with an mIoU of 61.19\%. 

The ablation results on the ISIC2018 dataset with lesions of different sizes (Small (S), Medium (M), and Large (L)) are shown in Table \ref{table_AblatiISIC2018}. The baseline model performs relatively poorly on small lesions (S), with an mIoU of only 64.04\%. EnCFBlock improves this to 68.51\%, while TCMA and CL achieve an mIoU of 67.97\% and 66.73\%, respectively. The combination of TCMA and EnCFBlock (+TCMA+EnCF) significantly improves the performance to 74.89\%. The full model achieves the best result for small lesions with an mIoU of 77.13\%, demonstrating that integrating all components enhances the segmentation accuracy and robustness.

\begin{table}[t]
\centering
\setlength{\tabcolsep}{3pt} 
\renewcommand{\arraystretch}{1.1} 
\fontsize{9pt}{9pt}\selectfont
\begin{tabular}{c c c c c c c}
    \hline
    \textbf{Methods} & \textbf{Size} & \textbf{MIOU} &  \textbf{DSC} &  \textbf{ACC} &  \textbf{SPE} &  \textbf{SEN} \\
    \hline
    
    \multirow{3}{*}{Baseline} & S & { 64.04} & { 78.07} & { 98.21} & { 98.37} & { 93.67} \\
    & M & { 78.90} & { 88.21} & { 97.03} & { 97.76} & { 91.75} \\
    & L & { 84.67} & { 91.70} & { 93.07} & { 95.42} & { 89.91} \\
    \hline
    \multirow{3}{*}{+EnCF} & S & { 68.51} & { 81.32} & { 98.54} & { 98.74} & { 93.04} \\
    & M & { 82.14} & { 90.19} & { 97.61} & { 98.55} & { 90.79} \\
    & L & { 84.71} & { 91.72} & { 93.05} & { 94.88} & { 90.57} \\
    \cline{2-7}
    \multirow{3}{*}{+TCMA} & S & { 67.97} & { 80.93} & { 98.49} & { 98.65} & { 94.02} \\
    & M & { 81.70} & { 89.93} & { 97.70} & { 98.26} & { 92.03} \\
    & L & { 84.62} & { 91.67} & { 92.98} & { 94.63} & { 90.76} \\
    \cline{2-7}
    \multirow{3}{*}{+CL} & S & { 66.73} & { 80.05} & { 98.43} & { 98.64} & { 92.44} \\
    & M & { 81.47} & { 89.79} & { 97.52} & { 98.57} & { 89.93} \\
    & L & { 85.60} & { 92.24} & { 93.51} & { 95.64} & { 90.64} \\
    \hline
    \multirow{3}{*}{\small +TCMA+EnCF} & S & { 74.89} & { 85.64} & { 98.95} & { 98.18} & { 92.33} \\
    & M & { 81.71} & { 89.94} & { 97.56} & { 98.58} & { 90.13} \\
    & L & { 85.41} & { 92.13} & { 93.46} & { 96.00} & { 90.02} \\
    \cline{2-7}
    \multirow{3}{*}{+EnCF+CL} & S & { 70.02} & { 82.36} & { 98.63} & { 98.79} & { 94.04} \\
    & M & { 81.74} & { 89.95} & { 97.51} & { 98.27} & { 92.02} \\
    & L & { 85.43} & { 92.14} & { 93.40} & { 95.16} & { 91.02} \\
    \cline{2-7}
    \multirow{3}{*}{+TCMA+CL} & S & { 72.23} & { 83.87} & { 98.76} & { 98.89} & { 94.96} \\
    & M & { 80.58} & { 89.25} & { 97.31} & { 98.00} & { 92.26} \\
    & L & { 83.75} & { 91.15} & { 92.68} & { 95.67} & { 88.65} \\
    \hline
    \multirow{3}{*}{S$^3$-Mamba} & S & { 77.13} & { 87.09} & { 99.07} & { 99.34} & { 91.66} \\
    & M & { 81.36} & { 89.72} & { 97.55} & { 98.81} & { 88.39} \\
    & L & { 83.28} & { 90.87} & { 92.60} & { 97.01} & { 86.65} \\
    \hline
\end{tabular}
\caption{Ablation results on the ISIC2018 dataset for Small (S), Medium (M) and Large (L) lesions.}
\label{table_AblatiISIC2018}
\end{table}

\begin{figure}[t]
\centering
\includegraphics[width=1\linewidth]{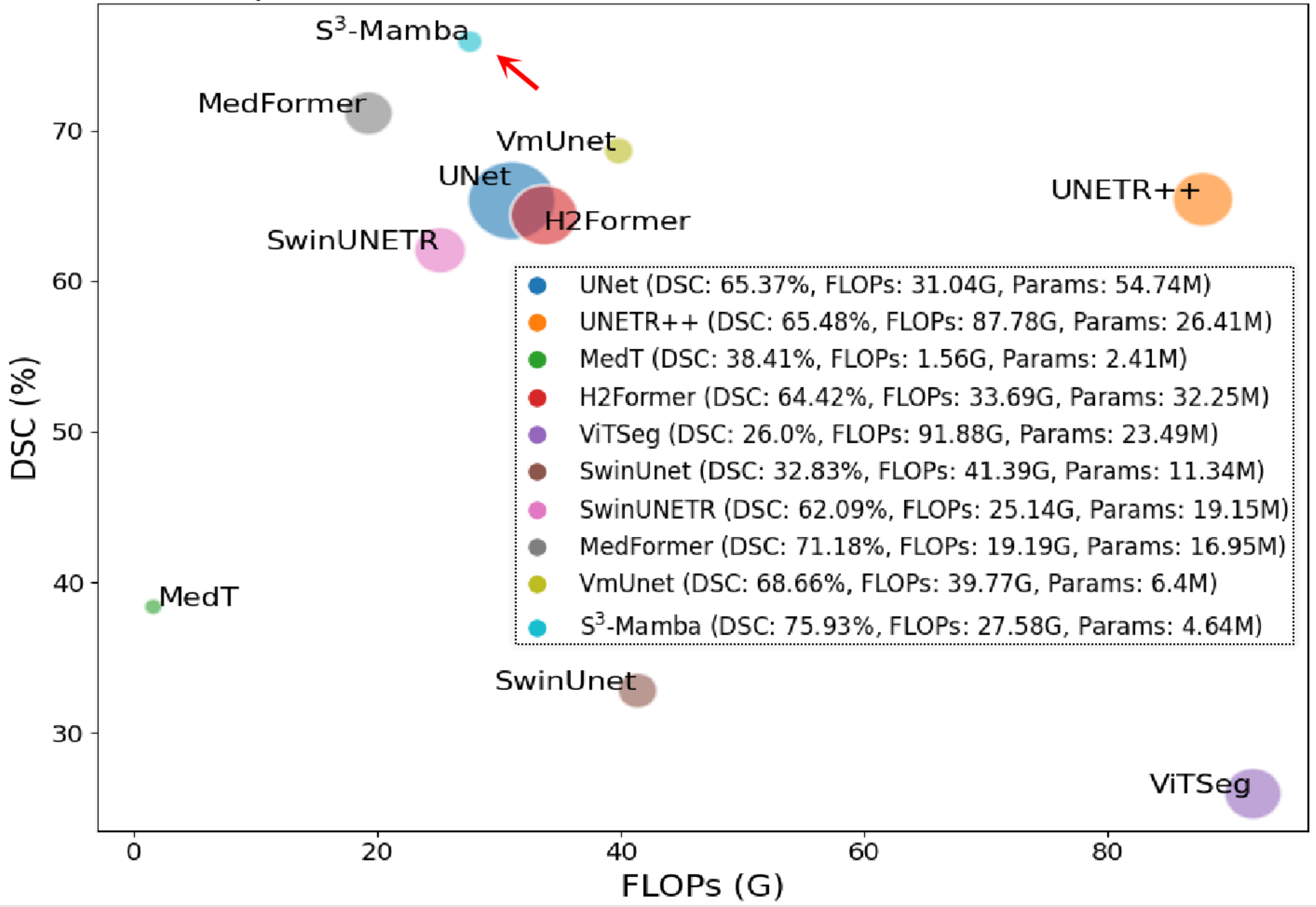} 
\caption{Comparison with other models in terms of FLOPs, DSC, and model size represented by the circle size on the Lymph dataset.} 
\label{figlcomples}
\end{figure}

\subsection{Model Complexity Analysis} 
Fig.~\ref{figlcomples} presents a scatter plot comparing various segmentation models based on FLOPs (Floating Point Operations per Second), DSC (Dice Similarity Coefficient), and Model Size (indicated by circle size). S$^3$-Mamba stands out as the most balanced model, achieving the highest DSC of 75.93\% with a moderate computational demand of 27.58G FLOPs and a compact model size of 4.64M parameters. It outperforms other models like UNETR++ (DSC 65.48\%, 87.78G FLOPs) and ViTSeg (DSC 26.0\%, 91.88G FLOPs), which are both computationally expensive and less accurate. While VmUnet also performs well, it requires more resources, \eg, 39.77G FLOPs and 6.4M parameters, while S$^3$-Mamba’s requires 27.58G FLOPs and 4.64M parameters. Both VmUnet and S$^3$-Mamba are based on the Mamba framework, showcasing the effectiveness of this architecture. Finally, our S$^3$-Mamba achieves a better segmentation accuracy of 75.93\% than 68.66\% for VmUnet while being more computationally efficient, making it the preferred choice for applications where both performance and efficiency are critical. 

\section{Conclusion}
In this paper, we propose Small-Size-Sensitive Mamba (S$^3$-Mamba), which significantly improves small lesion segmentation by introducing the EnVSSBlock, Tensor-based Cross-feature Multi-scale Attention, and a novel regularized curriculum learning strategy. In particular, the EnVSSBlock enhances the feature representation by adding residual connections to preserve local features and adaptively adjusting channel weights to amplify important features, particularly the subtle ones from small lesions through the channel-wise attention. The TCMA captures multi-scale information from patches of different modalities of different scales through a novel tensor-based attention mechanism, combining fine details with global context, thus improving the model's capability in segmenting small lesions. Finally, the novel regularized curriculum learning strategy enables the model to gradually learn features from large to small lesions, enhancing the segmentation accuracy and robustness. The combination of these strategies achieves significant performance gains over state-of-the-art models on three benchmark datasets, significantly enhancing small lesion segmentation. 

\section{Acknowledgments}
This work was supported by the National Natural Science Foundation of China under Grant 82261138629 and 12326610;  Guangdong Provincial Key Laboratory under Grant 2023B1212060076, and Shenzhen Municipal Science and Technology Innovation Council under Grant JCYJ20220531101412030.


\bibliography{aaai25}

\end{document}